\ifdefined\XeTeXversion\ifdefined\pdfminorversion\pdfminorversion=7\fi
\AddToHook{package/libertine/after}{\ifXeTeX
  \defaultfontfeatures[inconsolata]{Extension=.otf,
    UprightFont=Inconsolatazi4-Regular,BoldFont=Inconsolatazi4-Bold}
\fi}
\fi
\documentclass[acmsmall,nonacm,screen,10pt,balance=false]{acmart}
\newif\ifReportHTML
\IfFileExists{latexml.sty}{\RequirePackage{latexml}\iflatexml\ReportHTMLtrue\fi}{}
\ifReportHTML
\usepackage{amsmath,booktabs,array,tabularx,multirow}
\usepackage{graphicx,hyperref,natbib,enumitem}
\setcitestyle{authoryear,round,aysep={,}}
\newcommand{\model}{Antom 3-in-1 Transformer}
\newcommand{\merchantA}{Antom-Merchant A}
\newcommand{\merchantB}{Antom-Merchant B}
\newcommand{\pmstd}[1]{\ensuremath{\pm}#1}

\newcommand{\makecell}[2][]{\begingroup\def\\{\space}#2\endgroup}
\providecommand{\Description}[1]{}
\providecommand{\FloatBarrier}{}
\title{\model: A Payment Foundation Model}
\newcommand{\ReportAntAffiliation}{\affiliation{\institution{Ant International}}}
\newcommand{\ReportAuthor}[1]{\author{#1}\ReportAntAffiliation}
\newcommand{\ReportIntern}[2]{\author{#1}\affiliation{\institution{Ant International} (intern); \institution{#2}}}
\ReportAuthor{Weiyu Han}
\ReportAuthor{Haoyu Jiang}
\ReportAuthor{Jie Sheng}
\ReportAuthor{Youxi Pan}
\ReportAuthor{Xianglong Liu}
\ReportAuthor{Zeqi Ren}
\ReportAuthor{Dongquan Yi}
\ReportAuthor{Yinhao Liu}
\ReportIntern{Ruobing Zuo}{East China Normal University}
\ReportIntern{Ze Zhao}{Shanghai Jiao Tong University}
\ReportIntern{Huilin He}{Tongji University}
\ReportAuthor{Nairong Zhang}
\ReportAuthor{Xinyi He}
\author{Xiao Tan}\affiliation{\institution{Ant International}; corresponding author}\email{alex.tx@ant-intl.com}
\ReportAuthor{Jun Wang}
\ReportAuthor{Pan Fang}
\ReportAuthor{Wei Zeng}
\ReportAuthor{Jianan Su}
\ReportAuthor{Kaixin Shen}
\ReportAuthor{Jingtong Wu}
\ReportAuthor{Hao Li}
\ReportAuthor{Shuai Chen}
\ReportAuthor{Jiangming Yang}
\ReportAuthor{Tianyi Zhang}
\date{November 2026}

\else
\usepackage{fontawesome5}
\usepackage{amsmath,booktabs,array,tabularx,multirow,makecell}
\usepackage{placeins,microtype}
\usepackage{balance}
\usepackage{enumitem}
\IfFileExists{flafter.sty}{\usepackage{flafter}}{}
\usepackage{tikz}
\usetikzlibrary{arrows.meta,decorations.pathreplacing,calc}
\setcopyright{none}
\acmConference{}{}{}
\acmBooktitle{}
\acmDOI{}
\acmISBN{}
\setcitestyle{round,aysep={,},citesep={;},yysep={,}}
\title{\model{}}
\subtitle{A Payment\\Foundation Model}
\author{Weiyu Han, Haoyu Jiang, Jie Sheng, Youxi Pan, Xianglong Liu, Zeqi Ren, Dongquan Yi, Yinhao Liu, Ruobing Zuo, Ze Zhao, Huilin He, Nairong Zhang, Xinyi He, Xiao Tan, Jun Wang, Pan Fang, Wei Zeng, Jianan Su, Kaixin Shen, Jingtong Wu, Hao Li, Shuai Chen, Jiangming Yang, Tianyi Zhang}
\renewcommand{\shortauthors}{}
\date{}
\hypersetup{pdftitle={Antom 3-in-1 Transformer: A Payment Foundation Model},pdfauthor={Weiyu Han, Haoyu Jiang, Jie Sheng, Youxi Pan, Xianglong Liu, Zeqi Ren, Dongquan Yi, Yinhao Liu, Ruobing Zuo, Ze Zhao, Huilin He, Nairong Zhang, Xinyi He, Xiao Tan, Jun Wang, Pan Fang, Wei Zeng, Jianan Su, Kaixin Shen, Jingtong Wu, Hao Li, Shuai Chen, Jiangming Yang, Tianyi Zhang},colorlinks=true,linkcolor=black,citecolor=black,urlcolor=black}
\newcommand{\model}{{\fontspec[BoldFont=LinBiolinum_RB.otf]{LinBiolinum_RB.otf}\textbf{Antom 3-in-1 Transformer}}}
\newcommand{\merchantA}{Antom-Merchant A}
\newcommand{\merchantB}{Antom-Merchant B}
\newcommand{\pmstd}[1]{{\scriptsize$\pm$#1}}
\newcommand{\incfig}[2][\linewidth]{\includegraphics[width=#1]{#2}}

\ifXeTeX\AtBeginDvi{}\fi
\newcommand{\ReportAuthorNames}{%
\mbox{Weiyu Han\textsuperscript{1}}, \mbox{Haoyu Jiang\textsuperscript{1}}, \mbox{Jie Sheng\textsuperscript{1}}, \mbox{Youxi Pan\textsuperscript{1}}, \mbox{Xianglong Liu\textsuperscript{1}}\par
\mbox{Zeqi Ren\textsuperscript{1}},
\mbox{Dongquan Yi\textsuperscript{1}},
\mbox{Yinhao Liu\textsuperscript{1}},
\mbox{Ruobing Zuo\textsuperscript{1,2,*}},
\mbox{Ze Zhao\textsuperscript{1,3,*}},
\mbox{Huilin He\textsuperscript{1,4,*}}\par
\mbox{Nairong Zhang\textsuperscript{1}},
\mbox{Xinyi He\textsuperscript{1}},
\mbox{Xiao Tan\textsuperscript{1,\textdagger}},
\mbox{Jun Wang\textsuperscript{1}},
\mbox{Pan Fang\textsuperscript{1}},
\mbox{Wei Zeng\textsuperscript{1}}\par
\mbox{Jianan Su\textsuperscript{1}}, \mbox{Kaixin Shen\textsuperscript{1}}, \mbox{Jingtong Wu\textsuperscript{1}}, \mbox{Hao Li\textsuperscript{1}},
\mbox{Shuai Chen\textsuperscript{1}},\par
\mbox{Jiangming Yang\textsuperscript{1}},
\mbox{Tianyi Zhang\textsuperscript{1}}\par}
\newcommand{\ReportAuthorAffiliations}{%
  \mbox{\textsuperscript{1}Ant International}\quad
  \mbox{\textsuperscript{2}East China Normal University}\par
  \mbox{\textsuperscript{3}Shanghai Jiao Tong University}\quad
  \mbox{\textsuperscript{4}Tongji University}\par
}
\newcommand{\ReportAuthorNotes}{%
  \textsuperscript{*}Interns at Ant International.\par
  \textsuperscript{\textdagger}Corresponding Author,\enspace
  \faEnvelope[regular]\enspace
  \href{mailto:alex.tx@ant-intl.com}{alex.tx@ant-intl.com}\par
}

\usepackage{colortbl}
\makeatletter
\def\pgf@sys@bp@correct#1{#1=0.99627#1}
\edef\ReportOriginalRoman{\rmdefault}
\makeatother
\definecolor{ReportInk}{HTML}{252A30}
\definecolor{ReportMuted}{HTML}{707070}
\definecolor{ReportRule}{HTML}{BDBDBD}
\definecolor{ReportAccent}{HTML}{0369FF}
\definecolor{ReportPanel}{HTML}{E6F6FF}
\definecolor{ReportAbstractInk}{HTML}{1C2B33}
\newlength{\ReportOriginalColumnWidth}
\newlength{\ReportOriginalTextWidth}
\setlist[itemize]{beginpenalty=10000}
\newsavebox{\ReportGraphicBox}
\newlength{\ReportRequestedGraphicWidth}
\newlength{\ReportGraphicPlacementOffset}
\renewcommand{\incfig}[2][\linewidth]{%
  \begingroup
  \setlength{\textwidth}{\ReportOriginalTextWidth}%
  \setlength{\columnwidth}{\ReportOriginalColumnWidth}%
  \setlength{\ReportRequestedGraphicWidth}{#1}%
  \setlength{\ReportGraphicPlacementOffset}{0pt}%
  \ifdim\ReportRequestedGraphicWidth>\linewidth
    \setlength{\ReportRequestedGraphicWidth}{166mm}%
    \setlength{\ReportGraphicPlacementOffset}{.207bp}%
  \fi
  \ifstrequal{#2}{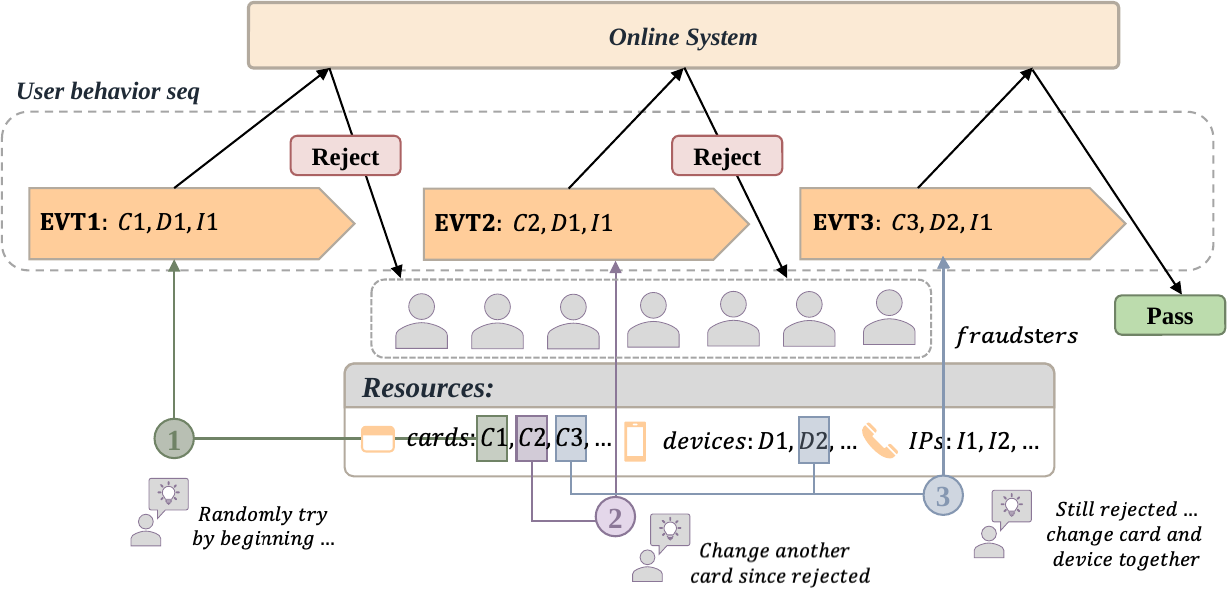}{\setlength{\ReportRequestedGraphicWidth}{123mm}}{}%
  \sbox{\ReportGraphicBox}{\includegraphics[width=\ReportRequestedGraphicWidth]{#2}}%
  \makebox[\linewidth][c]{\kern-\ReportGraphicPlacementOffset\usebox{\ReportGraphicBox}\kern\ReportGraphicPlacementOffset}%
  \endgroup}

\makeatletter
\renewcommand\normalsize{\@setfontsize\normalsize{11}{14}%
  \abovedisplayskip 9pt plus 2pt minus 2pt
  \belowdisplayskip 9pt plus 2pt minus 2pt
  \abovedisplayshortskip 3pt plus 2pt
  \belowdisplayshortskip 6pt plus 2pt minus 2pt}
\renewcommand\small{\@setfontsize\small{9}{11}}
\renewcommand\footnotesize{\@setfontsize\footnotesize{8}{10}}
\renewcommand\scriptsize{\@setfontsize\scriptsize{7}{8}}

\renewcommand\section{\@startsection{section}{1}{\z@}%
  {-16pt plus -2pt minus -1pt}{3.90328pt plus 1pt}%
  {\normalfont\sffamily\bfseries\fontsize{17}{21}\selectfont}}

\renewcommand\subsection{\@startsection{subsection}{2}{\z@}%
  {-13pt plus -2pt minus -1pt}{6pt plus 1pt}%
  {\normalfont\sffamily\bfseries\fontsize{14}{17}\selectfont}}

\renewcommand\subsubsection{\@startsection{subsubsection}{3}{\z@}%
  {-10pt plus -2pt minus -1pt}{5pt plus 1pt}%
  {\normalfont\sffamily\bfseries\fontsize{11}{14}\selectfont}}
\newsavebox{\ReportWideTableBox}
\newsavebox{\ReportCompactTableBox}
\renewenvironment{table}[1][!htbp]{%
  \@float{table}[#1]\centering
  \begin{lrbox}{\ReportCompactTableBox}\begin{minipage}{.8\linewidth}%
  \setlength{\textwidth}{\linewidth}\setlength{\columnwidth}{\linewidth}%
}{\end{minipage}\end{lrbox}\makebox[\linewidth][c]{\usebox{\ReportCompactTableBox}}\end@float}
\renewenvironment{table*}[1][!htbp]{%
  \@float{table}[#1]\centering
  \begin{lrbox}{\ReportWideTableBox}\begin{minipage}{\linewidth}%
  \setlength{\textwidth}{\linewidth}\setlength{\columnwidth}{\linewidth}%
}{\end{minipage}\end{lrbox}\makebox[\linewidth][c]{\usebox{\ReportWideTableBox}}\end@float}
\renewenvironment{figure*}[1][!htbp]{\@float{figure}[#1]}{\end@float}
\newsavebox{\ReportCardBox}
\newlength{\ReportCardHeight}
\renewcommand\maketitle{%
  \global\@ACM@maketitle@typesettrue
  \thispagestyle{firstpagestyle}
  \sbox{\ReportCardBox}{%
    \begin{minipage}[t]{\dimexpr\textwidth-34.016bp\relax}
      \parindent=0pt\parskip=0pt
{\sffamily\bfseries\fontsize{20.74}{23}\selectfont
 \hyphenpenalty=10000
 \exhyphenpenalty=10000
 \raisebox{-0.1em}{%
   \includegraphics[height=0.9em]{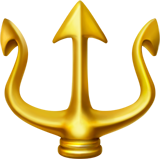}%
 }%
 \hspace{0.2em}%
 {\color{ReportAccent}\@title}%
 {\color{black}: \@subtitle}\par}
\vspace{9pt}
      {\sffamily\bfseries\fontsize{10}{13}\selectfont\color{black}\ReportAuthorNames}
      \vspace{6pt}
      {\normalfont\fontsize{9}{11.5}\selectfont\color{black}\microtypesetup{protrusion=false}\ReportAuthorAffiliations}
      \vspace{5pt}
      {\normalfont\fontsize{8.5}{10.5}\selectfont\color{black}\microtypesetup{protrusion=false}\ReportAuthorNotes}
      \vspace{14pt}
      {\normalfont\sffamily\bfseries\fontsize{11}{14}\selectfont\color{ReportAbstractInk}\abstractname\par}
      \vspace{5pt}
      {\normalfont\normalsize\color{ReportAbstractInk}\@abstract\par}
      \vspace{19pt}
      \noindent\begin{minipage}[c]{.62\linewidth}
        \sffamily\fontsize{9}{11.15}\selectfont\color{black}
        \textbf{Date:} November 2026
      \end{minipage}\hfill
      \begin{minipage}[c]{123pt}\raggedleft
        \includegraphics[width=123pt]{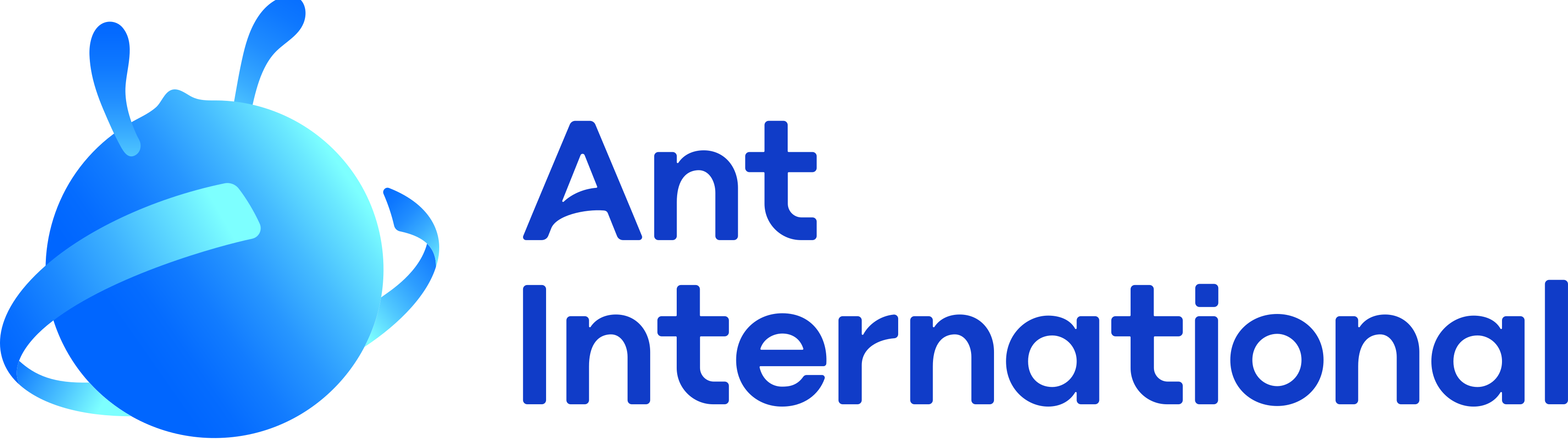}
      \end{minipage}
    \end{minipage}}
  \setlength{\ReportCardHeight}{\dimexpr\ht\ReportCardBox+\dp\ReportCardBox+41.262918bp\relax}%
  \ifdim\ReportCardHeight<431.5952bp\setlength{\ReportCardHeight}{431.5952bp}\fi
  \noindent\begin{tikzpicture}
    \path[use as bounding box] (0,0) rectangle (\textwidth,\ReportCardHeight);
    \fill[ReportPanel,rounded corners=10pt] (0,0) rectangle (\textwidth,\ReportCardHeight);
    \node[anchor=north west,inner sep=0pt] at (16.9313bp,\dimexpr\ReportCardHeight-24.254918bp\relax) {\usebox{\ReportCardBox}};
  \end{tikzpicture}\par
  \vspace{-9.7085pt}
}
\AtBeginDocument{%
  \normalsize\color{black}\raggedbottom
  \displaywidowpenalty=10000
  \fancypagestyle{standardpagestyle}{%
    \fancyhf{}%
    \fancyfoot[C]{\normalfont\normalsize\thepage}}
  \fancypagestyle{firstpagestyle}{%
    \fancyhf{}%
    \fancyfoot[C]{\normalfont\normalsize\thepage}}
  \pagestyle{standardpagestyle}}
\makeatother
\hypersetup{pdfauthor={Weiyu Han, Haoyu Jiang, Jie Sheng, Youxi Pan, Xianglong Liu, Zeqi Ren, Dongquan Yi, Yinhao Liu, Ruobing Zuo, Ze Zhao, Huilin He, Nairong Zhang, Xinyi He, Xiao Tan, Jun Wang, Pan Fang, Wei Zeng, Jianan Su, Kaixin Shen, Jingtong Wu, Hao Li, Shuai Chen, Jiangming Yang, Tianyi Zhang},pdftitle={Antom 3-in-1 Transformer: A Payment Foundation Model},
  colorlinks=true,linkcolor=black,citecolor=black,urlcolor=black}

\AtBeginDvi{}

\AtBeginDocument{%
  \hypersetup{
    colorlinks=true,
    citecolor=ReportAccent,
    linkcolor=black,
    urlcolor=black,
    filecolor=black
  }%
}

\fi

\begin{document}
\begin{abstract}

We report the development of \textbf{\model{}}, a payment foundation model that integrates \textbf{sequence}, \textbf{tabular}, and \textbf{graph} representations for payment risk detection. The model contains approximately 7B parameters, with approximately 1.7B activated per task route.
\model{} is pretrained to learn reusable behavioral and relational representations, primarily on Ant International's internal payment data.
In a cross-business transfer experiment, pretraining improves the target business metric by 34.5\% over the same architecture trained from scratch. With only 25\% of the training labels, the pretrained model outperforms the full-label scratch baseline in Average Precision (AP) and Recall@0.5\% on different merchant domains. In a four-month online experiment on Ant International's billion-scale payment platform, the model reduced the overall loss rate by 37.88\% relative to XGB-Online.

\end{abstract}

\maketitle
\hypertarget{introduction}{%
\section{Introduction}\label{introduction}}

Large-scale payment platforms need to continuously detect fraud across different markets and business scenarios. \textbf{Ant International} has accumulated substantial transaction data and experience in risk prevention as its business has expanded. \textbf{Antom}, a brand of Ant International, is a unified
merchant payment and digitalisation platform serving
businesses of all sizes across Asia and worldwide. User behavior, transaction processes, and data distributions differ across scenarios, while fraudsters continually adjust their tactics and create new forms of risk~\cite{he2026payseqgen}. These changes increase the work required to maintain effective risk systems across markets. They motivate a transaction foundation model that generalizes across scenarios and tactics and supports the continued prevention of complex payment risks.

Research on transaction foundation models has made substantial progress~\cite{zuo2026tapes,han2026tridentseq}. TransactionGPT learns transaction trajectories through autoregressive modeling and combines transaction metadata with downstream features \cite{ref1}. PRAGMA uses masked reconstruction to learn transferable representations from heterogeneous banking events and user attributes. PANTHER combines generative pretraining with contrastive learning and applies behavioral representations to business decisions \cite{ref3}. These approaches develop rich representations of individual histories. Payment risk also depends on relationships across accounts and payment objects. In its anti-money-laundering analysis, PRAGMA identifies the difficulty of capturing highly relational patterns through independent history encoding \cite{ref2}. The same transaction can also present different signals from different perspectives: an ordinary-looking user history may coexist with an associated network location or device that carries high-risk context. Jointly modeling these perspectives gives risk assessment a more complete transaction context~\cite{jiang2026interfuse}.

Based on these observations, we propose \model{}, a foundation model for card-payment risk that integrates transaction sequences, scenario-specific tabular features, and entity relationships. It has approximately 7B total parameters, with approximately 1.7B activated for a single task route. The model combines behavioral histories, transaction fields, and user relations that specialized risk models often process separately. Large-scale pretraining learns transaction behavior and entity relations; business adaptation connects this knowledge to the objectives and decision rules of each business.

The resulting foundation supports cross-scenario reuse, learning with limited labels, and behavioral analysis. In a transfer experiment, adapting an existing source-domain foundation to a new business improves the business metric by 34.5\% over the same architecture trained from scratch.
With only 25\% of the training labels, the pretrained
model outperforms the full-label scratch baseline
in Average Precision (AP) and Recall@0.5\% on both merchant domains.
 A shared-checkpoint study evaluates adaptation across Antom-Merchant A, Antom-Merchant B and AlipayHK. Pretrained representations also distinguish evasion and account-nurturing behavior. These studies examine the reusable behavioral capabilities contributed by pretraining. Joint-model comparisons then show how behavioral and relational information improves risk detection when combined with scenario features. In production, \model{} reduces overall loss rate by 37.88\% relative to XGB-Online at the same 0.5\% intervention budget.

\section{Scope and Limitations of this Technical Report}

\label{sec:scope-and-limitations}

This report focuses on the adaptation capabilities, label efficiency, and industrial deployment of \model{}. The model
is pretrained
through sequence contrastive learning and self-supervised learning for graph relationship, respectively,
primarily on Ant International's internal payment data. Subsequent supervised fine-tuning adapts the model to scenario-specific risk objectives across Antom and other Ant International businesses.

\begin{figure}[!t]
  \centering
  \ifReportHTML
  \includegraphics[width=\linewidth,alt={Sequence, tabular, and graph encoders feed task-conditioned attention and sparse experts; pretraining objectives learn behavioral and relational representations.}]{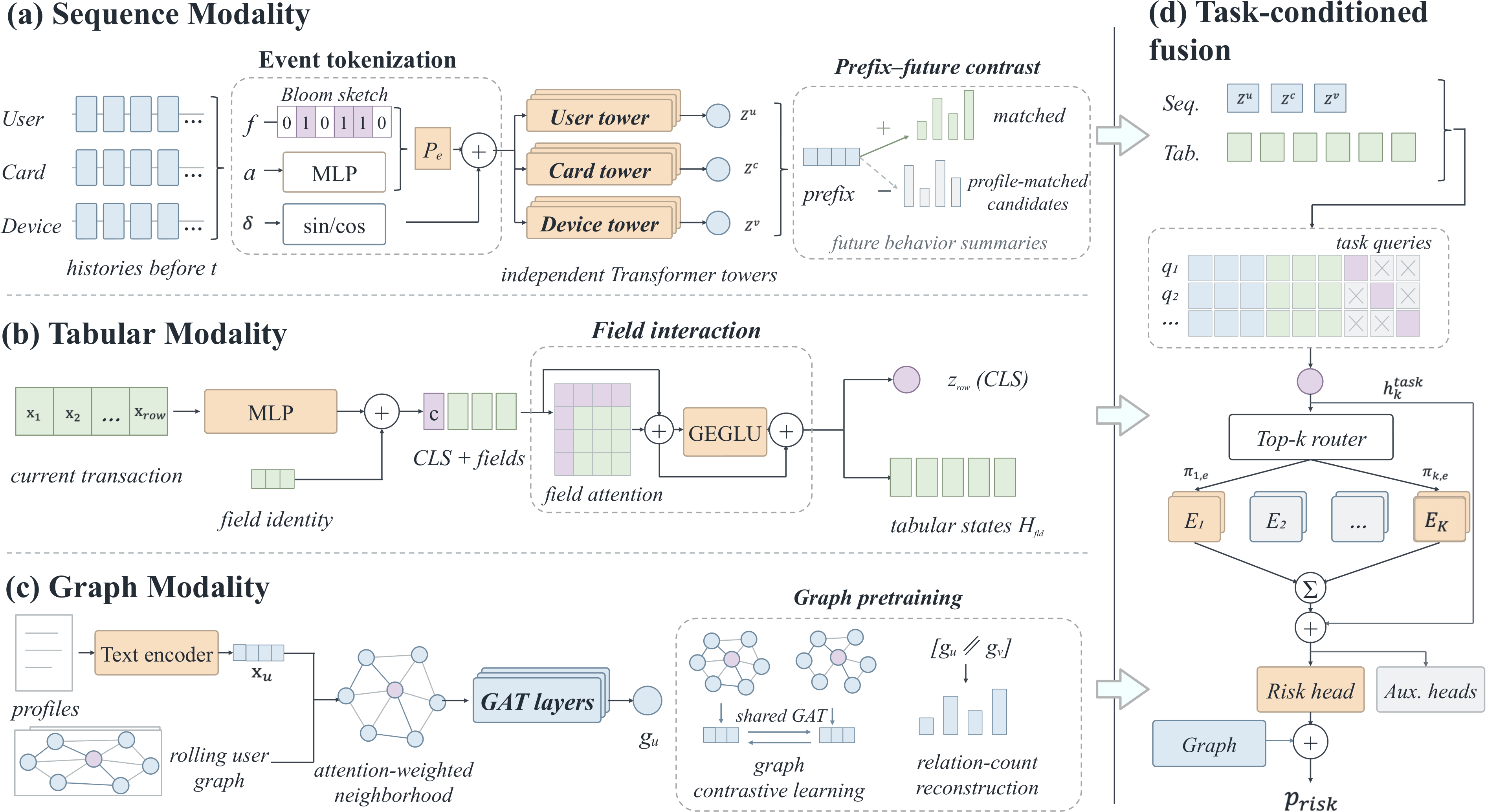}
  \else
  \includegraphics[width=\textwidth]{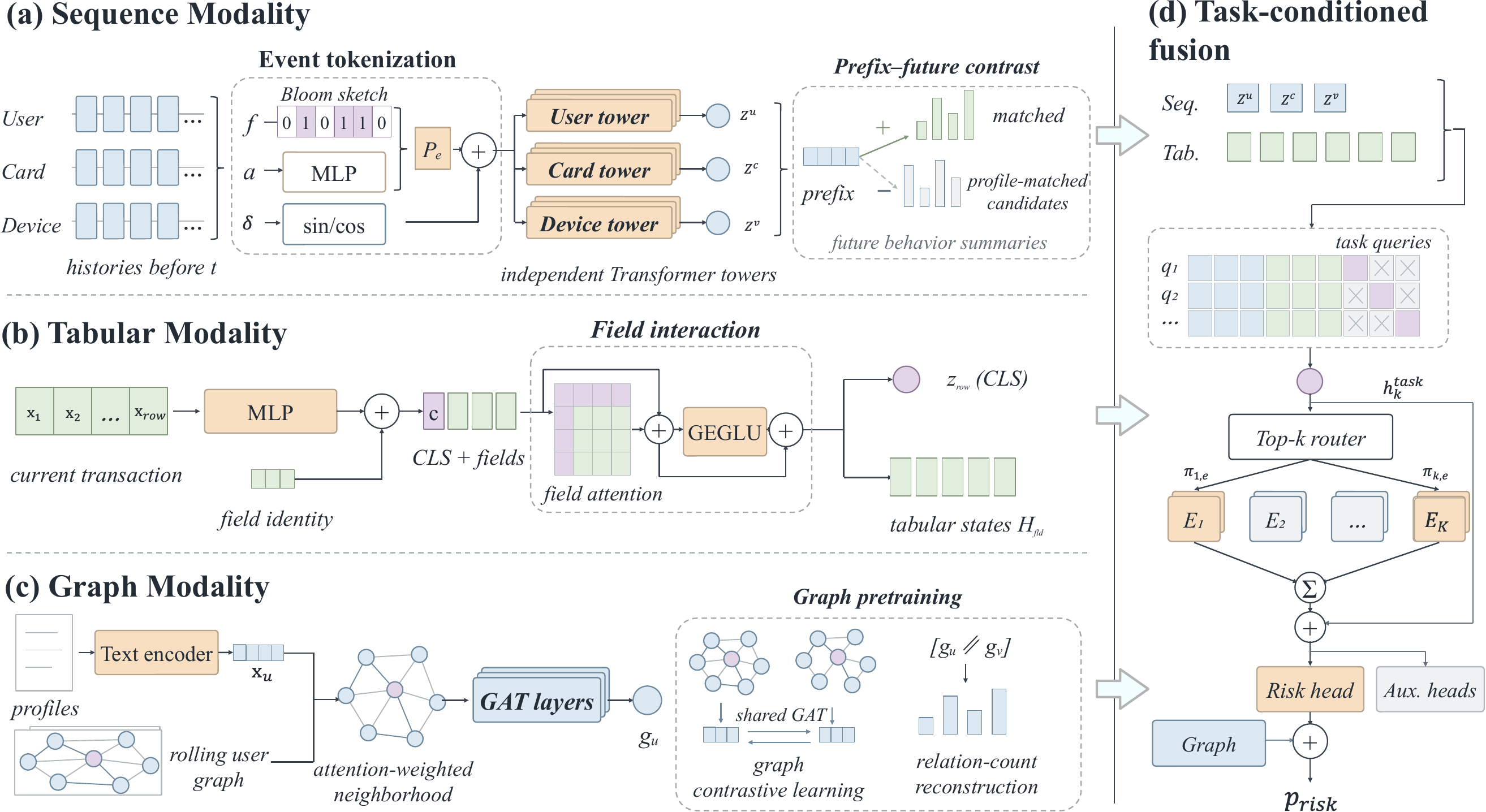}
  \fi
  \caption{
Overview of \textbf{\model{}}.
(a) \textbf{Sequence modality.}
Independent Transformer towers encode user, card, and device histories.
(b) \textbf{Tabular modality.}
Field attention captures interactions among transaction attributes and business features.
(c) \textbf{Graph modality.}
A graph attention network combines account profiles and rolling user
graphs to learn relational representations.
(d) \textbf{Task-conditioned fusion.}
Task queries and sparse experts fuse sequence and tabular representations;
graph embeddings augment the risk head.
}
  \Description{Sequence, tabular, and graph encoders feed task-conditioned attention and sparse experts; pretraining objectives learn behavioral and relational representations.}
  \label{fig:foundation-overview}
\end{figure}

We are committed to independent auditing of our technologies and have shared initial findings and selected methodological details on payment foundation models in research papers accompanying this release.
Detailed model architecture, training configurations, and dataset construction procedures are outside the scope of this release. We briefly introduce the model architecture and pretraining method here.
We plan to make further technical details available to additional third parties who can advise us on how to weigh the competitive and safety considerations above against the scientific value of further transparency.

\section{Methodology Overview}

We introduce \textbf{\model{}}, a payment foundation model
that integrates sequence, tabular, and graph representations for payment risk detection. Figure~\ref{fig:foundation-overview} presents the architecture of \model{}. The \textbf{sequence} modality learns behavioral patterns from transaction histories. The \textbf{tabular} modality models interactions
among transaction attributes, business states, and engineered
features through self-attention. The \textbf{graph} modality captures relationships among users by aggregating their profile and connected neighborhoods . Each modality has a corresponding encoder. Task-conditioned fusion combines sequence, tabular and graph representations for each prediction task.

\noindent\textbf{Sequence Encoder.}
The sequence encoder models the transaction histories associated with the current payment. \textit{Event tokenizer} embeds each event through its amount, categorical fields, and elapsed time, then uses a transformer to summarize the history. User, card, and device histories follow this procedure with separate encoders. The sequence encoders are pretrained by \textit{prefix-future contrast}, which matches each historical prefix with a statistical summary of its subsequent activity. Contrastive candidates with similar coarse activity profiles encourage the encoder to learn finer behavioral differences without risk labels.

\par\noindent\textbf{Tabular Encoder.}
The tabular encoder embeds features commonly used in payment-risk assessment, including current-transaction attributes and business statistics. The modeling procedure of tabular modality includes two stages: \textit{field tokenization} embeds each value, and \textit{field interaction} uses self-attention to combine information across fields. The resulting representations describe both individual fields and the transaction as a whole.

\par\noindent\textbf{Graph Encoder.}
The graph encoder models users through their account profiles and their connections to other users. We construct rolling user graphs from transactions in dynamic time interval. Each user node initializes from an embedding of its account.
The graph encoder is pretrained by combining contrastive agreement between edge-perturbed views of the same user with reconstruction of relation counts. The objectives preserve neighborhood consistency and relational information without supervision.

\par\noindent\textbf{Task-Conditioned Fusion.}
For task-conditioned fusion,
the model is optimized for a payment-risk prediction task together with a set of auxiliary business tasks. We associate each task with a learnable vector, referred to as a task query.
Learnable task queries aggregate information
through shared attention, and the resulting task representations
are used for expert routing.
Downstream fine-tuning adapts \model{} to supervised risk and auxiliary tasks. It updates the sequence encoder and trains the tabular encoder, task-conditioned fusion module, experts, and prediction heads. The graph encoder is frozen and supplies user representations aligned with each transaction's time.

\section{Experiments}
\label{sec:evaluation}

We evaluate \model{} through overall risk-prediction performance, production deployment, and reuse across businesses and tasks. We then examine sequence pretraining, graph pretraining,  task-conditioned fusion, followed by modality ablations and behavioral case studies.

\subsection{Experimental Setup}
\label{sec:experimental-setup}
\label{sec:eval-protocol}
\label{sec:seq-setup}
\label{sec:seq-coverage}

\textbf{Datasets.} We use five groups of datasets as follows.
\begin{itemize}[leftmargin=*,itemsep=1pt,topsep=2pt,parsep=0pt]
\item \textit{Industrial datasets.} We use sampled card-payment transactions from
multiple business scenarios at Ant International. Each transaction has tabular features, user/card/device sequences, and user graph information.
\item \textit{IBM dataset.} IBM's synthetic credit-card dataset~\cite{padhi2021tabformer} contains 24,386,900 transactions.
\item \textit{Low-label and cross-scenario datasets.} These card-payment transaction datasets are sampled from core business of Ant International, including Antom-Merchant A, Antom-Merchant B, and AlipayHK (HK). Antom-Merchant A represents a global e-marketplace, while Antom-Merchant B is a pooled cohort of merchants spanning
e-commerce, digital entertainment, travel and aviation,
AI services, and other industries. AlipayHK is another important business for Ant International.
Table~\ref{tab:seq-data} summarizes the three evaluation domains.
\item \textit{Graph datasets.} Amazon and YelpChi are multi-relation review graphs~\cite{dou2020caregnn}, T-Finance is a financial transaction graph~\cite{tang2022bwgnn}, and Elliptic is a Bitcoin transaction-flow graph~\cite{weber2019elliptic}. Their statistics appear in Table~\ref{tab:graph-protocol}.
\item \textit{Fusion datasets.} We use Sampled Industrial, TransXion, IBM CCT, and MBD. Appendix~\ref{app:datasets} describes their inputs and splits (Table~\ref{tab:fusion-data}).
\end{itemize}

\textbf{Baselines.} Online risk prediction compares \model{} with XGB-Online~\cite{chen2016xgboost} and the SAINT-style production model~\cite{somepalli2021saint}. Public transaction baselines include FATA-Trans, TabBERT-LSTM,
PANTHER, TransactionGPT, and PRAGMA~\cite{zhang2023fatatrans,padhi2021tabformer,ref1,ref2,ref3}.

Sequence pretraining compares PFC with CoLES, BYB, and CPC~\cite{babaev2022coles,wu2025byb,oord2018cpc}, alongside training from scratch.

Graph augmentation comparisons include tabular-only and graph-only models, concatenation, graph-token fusion, logit stacking, and residual predictors.

Fusion comparisons use a fixed model reference (denoted as B1), score-only fusion (denoted as B2), concatenation, sum-pooling, and learned fusion methods: CE, MMT, UMI, $I^2M^2$, MMPareto, $I^2$MoE, and BALM~\cite{liu2019ce,gabeur2020mmt,zhang2023umi,madaan2024i2m2,wei2024mmpareto,xin2025i2moe,nguyen2026balm}.

\textbf{Model and comparison settings.}
The industrial configuration contains approximately
6.92B parameters, with approximately 1.74B
activated per task route. Sequence pretraining
objectives are compared using a common encoder,
pretraining data, and training budget. Graph and
fusion comparisons follow their respective matched
evaluation protocols.

\textbf{Metrics.} We evaluate ranking quality offline with three metrics and business impact online with two loss rates. Average precision (AP) summarizes precision over the risk ranking. AUROC measures class separation across score thresholds. Top-budget recall R@$q$ is the share of risk events captured among the highest-scoring $q$ fraction of transactions; we emphasize R@0.5\%, the production intervention budget. Graph benchmarks report AUPRC~\cite{tang2023gadbench}. For online assessment, the amount-based chargeback (CB) rate divides chargeback amount by successful transaction amount. The overall loss rate uses the same denominator and counts the realized financial loss after chargeback-dispute recoveries. Appendix~\ref{app:metrics} defines supplementary metrics.

\subsection{Performance Overview}
\label{sec:eval-public}

We assess overall performance in two settings: \textit{Industrial} online A/B test against the previous risk models; \textit{Open-source} comparison with other pretrained models and transaction foundation models.

\textbf{Industrial Online A/B test.}
\label{sec:eval-online}
From June 8 to October 14, 2026, the online evaluation covered 30\% of platform transactions. Transactions were randomly assigned to \model{} and two production models: XGB-Online, which periodically refits gradient-boosted trees, and the SAINT-style model. The three arms used the same business policies and intervention budget. Across the evaluated payment businesses, \model{} reduces the pooled CB rate by 37.74\% and the overall loss rate by 37.88\% relative to XGB-Online (Table~\ref{tab:online-ab}). Relative to SAINT-style, the reductions are 28.26\% and 24.07\%.

\begin{table}[!htbp]
\caption{Online portfolio results relative to XGB-Online.}
\label{tab:online-ab}
\small\setlength{\tabcolsep}{2pt}
\begin{tabular*}{\columnwidth}{@{\extracolsep{\fill}}lrrr@{}}
\toprule
System & \makecell[r]{Traffic\\share} & \makecell[r]{CB rate\\change} & \makecell[r]{Loss rate\\change}\\
\midrule
XGB-Online & 32.1\% & base & base\\
SAINT-style & 37.4\% & $-$13.21\% & $-$18.18\%\\
\makecell[l]{\model{}} & 30.4\% & \textbf{$-$37.74\%} & \textbf{$-$37.88\%}\\

\bottomrule
\end{tabular*}
\end{table}

The online \model{} was evaluated in two stages. In Stage~I (June 8--August 2), all three arms scored traffic without intervening. Subsequent confirmed outcomes measure each arm's recall among its top 0.5\% of scores. In Stage~II (August 3--October 14), each arm made production decisions under the same policies and a 0.5\% intervention budget, and we compare the realized CB rates. Financial outcomes include records received through October 25, 2026.

In Stage~I, \model{} recalls 29.79\% of risk transactions, compared with 7.39\% for XGB-Online and 16.52\% for SAINT-style. In Stage~II, its CB rate is 28.28\% lower than that of XGB-Online; weighting the arms by a common distribution of transaction dates gives a 26.09\% reduction (Table~\ref{tab:online-stages}). Stronger risk ranking thus translates into lower realized losses under the deployed intervention policy. Appendix~\ref{app:online-stages} specifies the transaction-date aggregation.

\begin{table}[!htbp]
\caption{Two-stage card-payment evaluation. Stage~II reports CB-rate results relative to XGB-Online.}
\label{tab:online-stages}
\small\setlength{\tabcolsep}{2pt}
\begin{tabular*}{\columnwidth}{@{\extracolsep{\fill}}lrrr@{}}
\toprule
& Stage I & \multicolumn{2}{c}{Stage II}\\
\cmidrule(lr){3-4}
System & R@0.5\% & Pooled & Date-weighted\\
\midrule
XGB-Online & 7.39\% & base & base\\
SAINT-style & 16.52\% & -12.66\% & -12.49\%\\
\makecell[l]{\model{}} & \textbf{29.79\%} & \textbf{-28.28\%} & \textbf{-26.09\%}\\
\bottomrule
\end{tabular*}
\end{table}

\textbf{Open-source comparison results.} On the temporal test split of IBM’s synthetic credit-card dataset, we compare \model{} with baseline configurations from five published approaches (Table~\ref{tab:public-benchmark}). \model{} achieves the highest reported AP, AUROC, R@0.5\%, and R@1\% among the evaluated configurations. Its AP of 0.3829 exceeds PRAGMA’s 0.3412 by 12.2\% on a relative basis. At the 0.5\% and 1\% intervention budgets, \model{} recalls 83.22\% and 97.17\% of fraud, compared to 72.12\% and 80.67\% for PRAGMA, corresponding to increases of 11.10 and 16.50 percentage points. \model{} also attains the highest AUROC of 0.9966, compared with the five evaluated baselines.

\begin{table*}[!htbp]
\caption{\model{} and transaction-model baselines on the temporal test split of IBM's synthetic dataset. R@$q$ is fraud recall among the top-$q$ scored transactions.}
\label{tab:public-benchmark}
\small
\begin{tabular*}{\textwidth}{@{\extracolsep{\fill}}lrrrrrr@{}}
\toprule
Model & AP & AUROC & F1 & R@0.1\% & R@0.5\% & R@1\%\\
\midrule
PANTHER-Full~\cite{ref3} & 0.0745 & 0.9567 & 0.1357 & 0.1241 & 0.3190 & 0.4600\\
FATA-Trans~\cite{zhang2023fatatrans} & 0.1832 & 0.9521 & 0.2490 & 0.2419 & 0.4732 & 0.5852\\
TabBERT-LSTM~\cite{padhi2021tabformer} & 0.2524 & 0.9872 & 0.3439 & 0.3234 & 0.6219 & 0.7316\\
TransactionGPT~\cite{ref1} & 0.2157 & 0.9705 & 0.2746 & 0.2649 & 0.4814 & 0.5805\\
PRAGMA~\cite{ref2} & 0.3412 & 0.9865 & \textbf{0.4091} & \textbf{0.3773} & 0.7212 & 0.8067\\
\midrule
\textbf{\model{}} & \textbf{0.3829} & \textbf{0.9966} & 0.3838 & 0.3642 & \textbf{0.8322} & \textbf{0.9717}\\
\bottomrule
\end{tabular*}
\end{table*}

\subsection{Industrial Implementation}
\label{sec:industrial}
\label{sec:eval-efficiency}

\model{} runs inside Ant International's real-time risk-control workflow. Every card payment triggers one scoring request: the feature platform assembles the model inputs, a GPU model service computes the risk score, and the strategy platform turns the score into a decision. The sequence and tabular encoders, task-conditioned fusion, and experts run online for each request, whereas graph embeddings are computed offline and only looked up at scoring time.

\begin{figure}[!htbp]
\centering
\ifReportHTML
\includegraphics[width=\linewidth,alt={A horizontal time axis from the payment event to the risk decision, with the feature platform, the model service, and the strategy platform in sequence, their monitored latencies, and the 120 ms and 250 ms budgets.}]{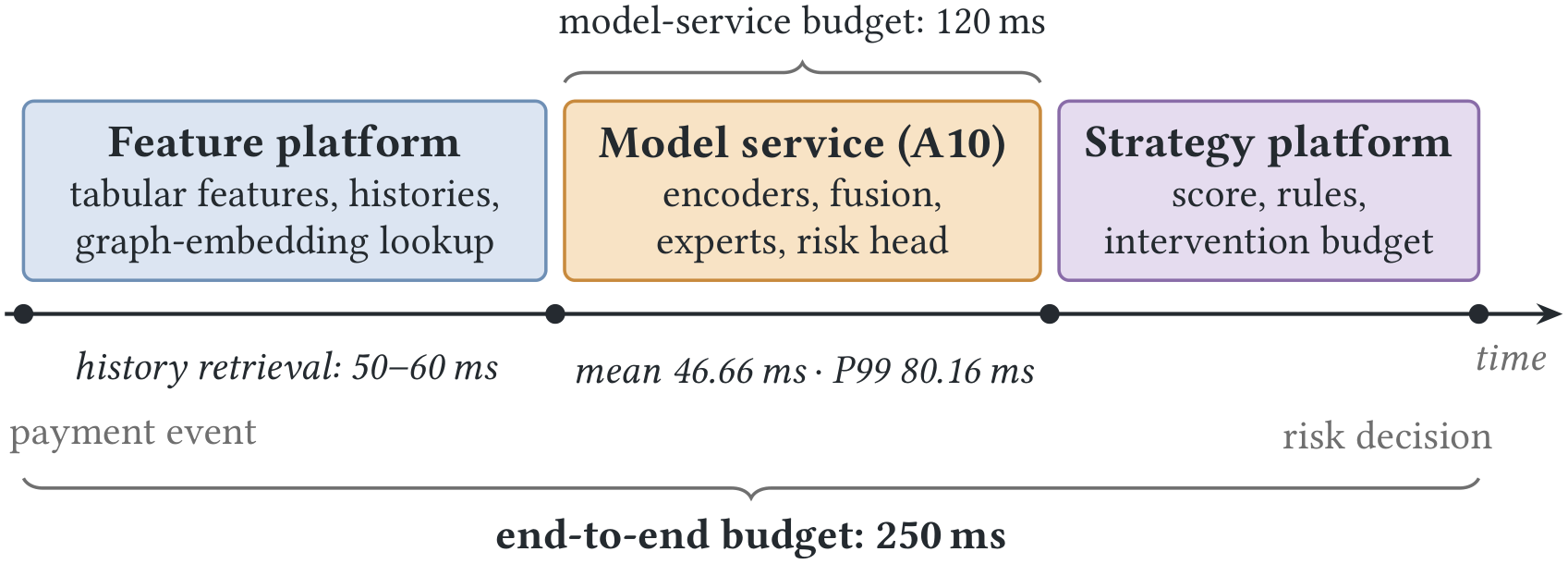}
\else
  \begingroup\fontencoding{TU}\fontfamily{\ReportOriginalRoman}\fontsize{10}{12}\selectfont
  \sbox{\ReportGraphicBox}{\resizebox{112mm}{!}{\input{figs_src/online_timeline}}}%
  \usebox{\ReportGraphicBox}\endgroup
\fi
\caption{Online scoring timeline of \model{}. Latencies are monitored values; braces mark service budgets. Stage widths are schematic.}
\Description{A horizontal time axis from the payment event to the risk decision, with the feature platform, the model service, and the strategy platform in sequence, their monitored latencies, and the 120 ms and 250 ms budgets.}
\label{fig:online-service}
\end{figure}

\textbf{Online system.} A payment event first enters the risk-control system. The feature platform computes the tabular features of the current transaction, retrieves the preceding user, card, and device histories, and
looks up the latest eligible graph embedding of the user.
History retrieval typically takes 50--60\,ms. The model service encodes the assembled inputs and returns a risk score. The strategy platform combines the score with the active rules and intervention budget and returns the risk decision. The complete workflow must finish within a 250\,ms end-to-end budget, which covers feature access, model-service communication and inference, and the final decision (Figure~\ref{fig:online-service}).

\textbf{Inference performance.}
The model service runs on an NVIDIA A10 GPU
under a 120\,ms service budget.
Online monitoring records a mean latency
of 46.66\,ms, a P95 of 55.86\,ms,
and a P99 of 80.16\,ms.

\subsection{Foundation Model Capabilities}
\label{sec:seq-transfer}

We examine how a pretrained foundation reduces the labels needed for a new business, transfers to other business scenarios and explore how the foundation model could benefit other tasks.

\begin{figure}[!htbp]
\centering
  \ifReportHTML
  \includegraphics[width=\linewidth,alt={Four plots show AP and recall at a 0.5 percent intervention budget for the pretrained model and scratch training across five training-label budgets on AliExpress and ANTOM, with a dashed line for full-label scratch.}]{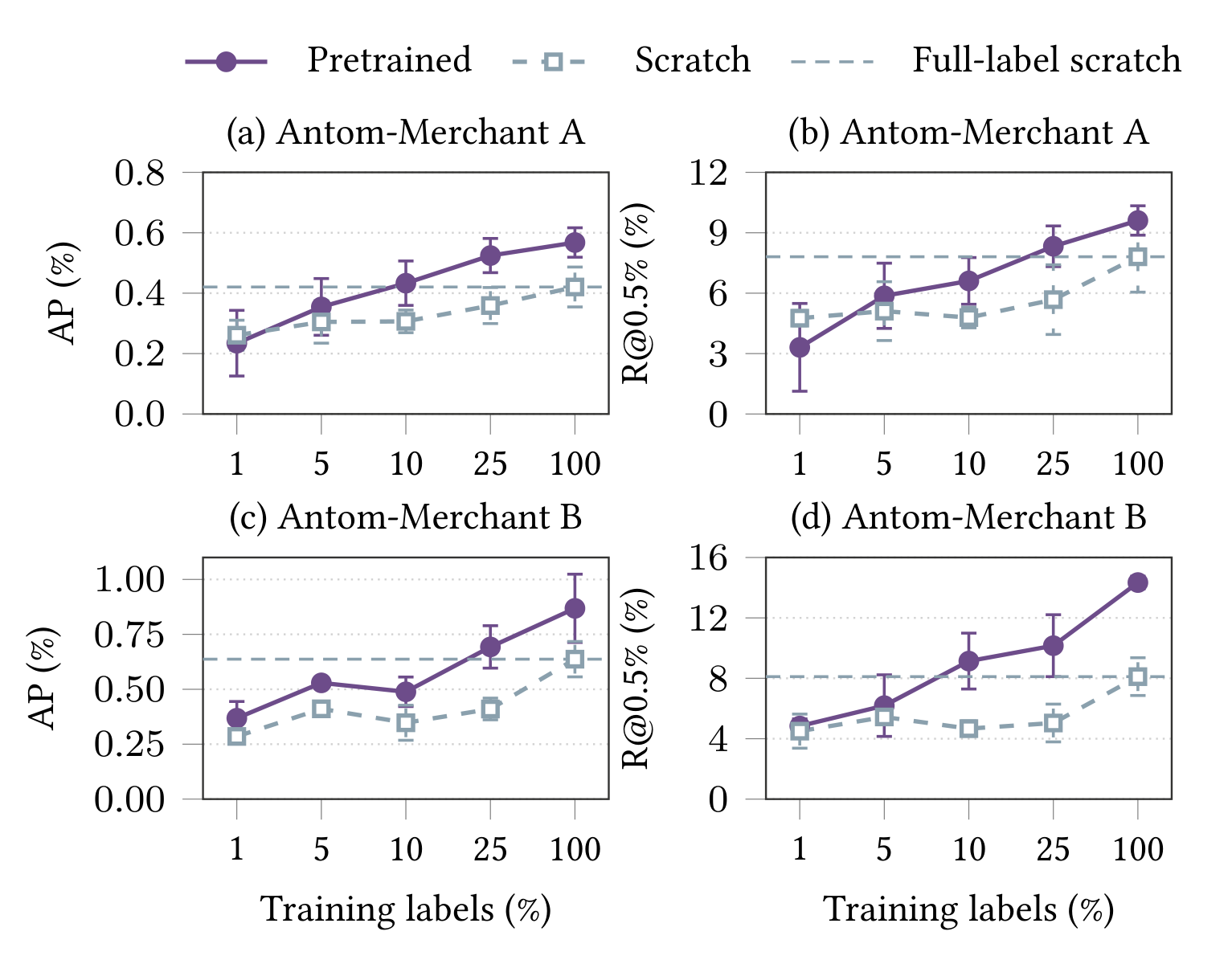}
  \else
  \incfig[\columnwidth]{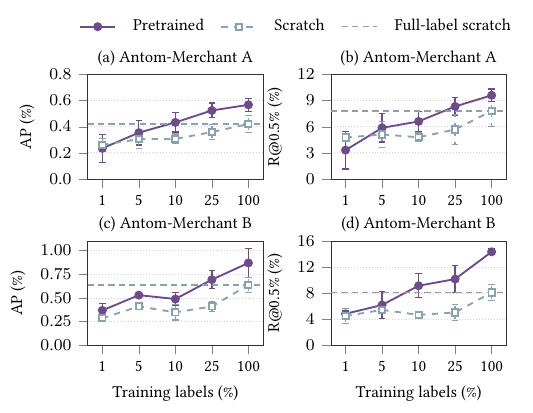}
  \fi

\caption{Label efficiency on \merchantA{} and
\merchantB{}. Points and error bars show the
mean and standard deviation over three
fine-tuning runs.}

  \Description{Four plots show AP and recall at a 0.5 percent intervention budget for the pretrained model and scratch training across five training-label budgets on AliExpress and ANTOM, with a dashed line for full-label scratch.}
\label{fig:label-efficiency}
\end{figure}

\textbf{Low-label adaptation.}
To evaluate label efficiency, we fine-tune the pretrained model and the same architecture trained from scratch with 1\%, 5\%, 10\%, 25\%, and 100\% of the training labels, keeping the validation set fixed.
With 25\% of the labels, the pretrained model already exceeds scratch training with all
labels on both \merchantA{} and \merchantB{} (Figure~\ref{fig:label-efficiency}).
On \merchantA{}, it reaches 0.5246\% AP and 8.33\%
R@0.5\%, against 0.4205\% and 7.81\% for full-label
scratch; on \merchantB{}, it reaches 0.6928\% and
10.16\%, against 0.6370\% and 8.11\%.
Pretraining also flattens the label curve: with 25\%
of the labels, the pretrained model retains 87\%
of its full-label recall on \merchantA{} and 71\%
on \merchantB{}, compared with 73\% and 62\%
for scratch training.

HK is the most label-scarce business, with 732 positive training examples, so we evaluate it at 10\% and 100\% of the labels only. With 10\% of the labels, the pretrained model reaches 0.8214\% AP and 27.75\% R@0.5\%, well above scratch training with all labels (0.1148\% and 10.41\%).
Across the evaluated budgets, pretraining improves
both metrics over same-budget scratch on HK
and \merchantB{}.

\textbf{Cross-scenario application.}
To evaluate the cross-scenario capability, we first fine-tune a pretrained model on a target business whose histories were excluded from pretraining. The pretrained model and its scratch counterpart share the architecture, target labels, fine-tuning budget, and selection rule. Pretraining improves the target business metric by 34.5\% and AUROC by 0.0518 over scratch. The gain does not depend on a few risk-dense accounts: after removing the accounts with the highest concentration of risk events, the business-metric gain remains 34.2\%.

We then test how the pretraining data source affects transfer. Three checkpoints of the same size and budget are pretrained on \merchantA{} histories only, on \merchantB{} histories only, and on an equal mixture of both. Each checkpoint is fine-tuned on each target business with 10\% or 25\% of its labels (Table~\ref{tab:cross-source}). Pretraining only on the other business already improves AP and R@0.5\% over scratch in all four target--budget settings. For \merchantB{} with 10\% labels, \merchantA{} histories help even more than \merchantB{}'s own (0.5791\% vs.\ 0.4857\% AP; 8.67\% vs.\ 6.13\% R@0.5\%). Histories in the \merchantB{} cohort
are shorter on average, with 7.8 events compared
with 29.5 for \merchantA{}. Thus, richer histories from another scenario can partly substitute for in-domain ones. With 25\% labels, mixing the two sources beats target-only pretraining on both metrics for both targets.

\begin{table}[!htbp]
\caption{Cross-scenario transfer by pretraining source
(\%). Bold: best per target domain and label budget.}
\label{tab:cross-source}
\small\setlength{\tabcolsep}{2pt}
\begin{tabular*}{\columnwidth}{@{\extracolsep{\fill}}lrrrr@{}}
\toprule
& \multicolumn{2}{c}{10\% labels}
& \multicolumn{2}{c}{25\% labels} \\
Pretraining source & AP & R@0.5\%
                   & AP & R@0.5\% \\
\midrule
\multicolumn{5}{l}{Target: \merchantA{}} \\
Scratch & 0.3067 & 4.79 & 0.3591 & 5.68 \\
\merchantA{} only
    & \textbf{0.3994} & 5.97 & 0.4361 & 7.85 \\
\merchantB{} only
    & 0.3668 & 5.96 & \textbf{0.4663} & \textbf{8.70} \\
\merchantA{} + \merchantB{}
    & 0.3331 & \textbf{6.46} & 0.4511 & 7.98 \\
\midrule
\multicolumn{5}{l}{Target: \merchantB{}} \\
Scratch & 0.3476 & 4.67 & 0.4103 & 5.04 \\
\merchantB{} only
    & 0.4857 & 6.13 & 0.5095 & 7.86 \\
\merchantA{} only
    & \textbf{0.5791} & \textbf{8.67} & 0.4319 & 6.94 \\
\merchantA{} + \merchantB{}
    & 0.4315 & 7.15 & \textbf{0.6379} & \textbf{9.05} \\
\bottomrule
\end{tabular*}
\end{table}

\textbf{Payment routing task application.}
Besides risk assessment, \model{} serves payment routing tasks. The payment routing tasks' purpose is to route each payment to suitable channels and improve the payment success rate (\textbf{PSR}). By post-training on payment routing tasks, we found that \model{} improves PSR by 2.16\% relative to our internal deployed model. \model{} allows requests containing both risk and payment routing features to detect transaction risk and improve PSR.

\subsection{Modality Pretraining and Fusion Results}
\label{sec:modality-results}

We evaluate sequence pretraining and two main components of task-conditioned fusion block: graph-augmented prediction and task-attention based expert fusion.

\textbf{Sequence pretraining.}
\label{sec:pfc-design-study}
We compare PFC with CoLES, BYB, and CPC using the same encoder, pretraining data, and training budget (Table~\ref{tab:seq-main}). \textbf{PFC achieves the best AP and R@0.5\% across all three evaluation domains at both label budgets.} With 10\% of the HK labels, it reaches 0.8214\% AP and 27.75\% R@0.5\%, compared with 0.0342\% and 3.31\% for scratch training.

\begin{table*}[!htbp]
\caption{Sequence pretraining objectives across evaluation domains and label budgets (\%). Mean $\pm$ standard deviation over three fine-tuning runs. Bold: best; underline: second best.}
\label{tab:seq-main}
\small
\setlength{\tabcolsep}{3pt}
\begin{tabular*}{\textwidth}{@{\extracolsep{\fill}}lrrrrrr@{}}
\toprule
& \multicolumn{2}{c}{\merchantA{}} & \multicolumn{2}{c}{\merchantB{}} & \multicolumn{2}{c}{HK}\\
\cmidrule(lr){2-3}\cmidrule(lr){4-5}\cmidrule(lr){6-7}
Model & AP & R@0.5\% & AP & R@0.5\% & AP & R@0.5\%\\
\midrule
\multicolumn{7}{@{}l}{\emph{10\% training labels}}\\
Scratch & 0.3067\pmstd{0.0375} & 4.79\pmstd{0.51} & 0.3476\pmstd{0.0798} & 4.67\pmstd{0.51} & 0.0342\pmstd{0.0082} & 3.31\pmstd{0.77}\\
CoLES (adapted) & 0.2879\pmstd{0.0182} & 5.18\pmstd{0.84} & 0.3514\pmstd{0.0691} & 5.04\pmstd{0.52} & 0.0274\pmstd{0.0124} & 2.92\pmstd{0.55}\\
BYB (adapted) & 0.2447\pmstd{0.0574} & 3.56\pmstd{1.80} & 0.3624\pmstd{0.0488} & \underline{5.91}\pmstd{2.17} & \underline{0.0475}\pmstd{0.0160} & 2.86\pmstd{0.56}\\
CPC (adapted) & \underline{0.3569}\pmstd{0.0094} & \underline{5.79}\pmstd{0.70} & \underline{0.3874}\pmstd{0.0283} & 5.60\pmstd{0.68} & 0.0279\pmstd{0.0061} & \underline{3.34}\pmstd{2.14}\\
\textbf{PFC (ours)} & \textbf{0.4331}\pmstd{0.0735} & \textbf{6.61}\pmstd{1.16} & \textbf{0.4886}\pmstd{0.0672} & \textbf{9.14}\pmstd{1.85} & \textbf{0.8214}\pmstd{0.3337} & \textbf{27.75}\pmstd{5.44}\\
\midrule
\multicolumn{7}{@{}l}{\emph{100\% training labels}}\\
Scratch & 0.4205\pmstd{0.0660} & 7.81\pmstd{1.76} & 0.6370\pmstd{0.0805} & 8.11\pmstd{1.25} & 0.1148\pmstd{0.0486} & 10.41\pmstd{5.19}\\
CoLES (adapted) & 0.4919\pmstd{0.0605} & 8.25\pmstd{0.77} & 0.3865\pmstd{0.0268} & 6.10\pmstd{0.84} & 0.1150\pmstd{0.1515} & 5.87\pmstd{3.23}\\
BYB (adapted) & \underline{0.5087}\pmstd{0.0211} & \underline{9.24}\pmstd{0.34} & \underline{0.7682}\pmstd{0.2357} & 9.80\pmstd{1.47} & 0.0525\pmstd{0.0199} & 6.76\pmstd{2.12}\\
CPC (adapted) & 0.5057\pmstd{0.0497} & 7.48\pmstd{3.47} & 0.7678\pmstd{0.1896} & \underline{10.89}\pmstd{0.10} & \underline{0.3128}\pmstd{0.0968} & \underline{12.74}\pmstd{2.74}\\
\textbf{PFC (ours)} & \textbf{0.5678}\pmstd{0.0487} & \textbf{9.61}\pmstd{0.73} & \textbf{0.8685}\pmstd{0.1553} & \textbf{14.34}\pmstd{0.42} & \textbf{0.8355}\pmstd{0.4142} & \textbf{28.33}\pmstd{3.25}\\
\bottomrule
\vspace{2mm}
\end{tabular*}
\end{table*}

\textbf{Graph-augmented prediction.}
\label{sec:graph-evaluation}
We compare our graph--augmented prediction method with other standard components on four public benchmarks(Table~\ref{tab:graph-public}). Our method applies a fixed pretrained graph encoder and adds a graph residual to the risk prediction. We denote our graph--augmented prediction method as Transaction Graph Reinforcement Fusion (TGRF). TGRF-Joint trains the backbone jointly with the fixed graph encoder, while TGRF-Frozen holds the backbone fixed. \textbf{TGRF-Joint achieves the best average rank across the four benchmarks.} It ranks first or second on each dataset, with an average rank of 1.75. On industrial payments, TGRF improves AUROC by 1.36\% and R@1\% by 7.11\% over the backbone.

\begin{table*}[!htbp]
\caption{Graph-augmented prediction on public graph benchmarks (AUPRC; mean $\pm$ standard deviation over ten seeds). Bold: best; underline: second best.}
\label{tab:graph-public}
\small
\begin{tabular*}{\textwidth}{@{\extracolsep{\fill}}lrrrrr@{}}
\toprule
Method & Amazon & YelpChi & T-Finance & Elliptic & Avg. rank $\downarrow$\\
\midrule
No graph (standalone) & .9184\pmstd{.0066} & .6553\pmstd{.0079} & .6782\pmstd{.0647} & .6578\pmstd{.0380} & 7.00\\
Graph only & .9032\pmstd{.0052} & .6579\pmstd{.0079} & .8341\pmstd{.0018} & .3129\pmstd{.0608} & 9.00\\
Concatenation (no dropout) & .9096\pmstd{.0088} & .7128\pmstd{.0113} & .8404\pmstd{.0056} & .6349\pmstd{.0320} & 6.00\\
Concatenation + dropout & .9110\pmstd{.0091} & .7107\pmstd{.0091} & .8411\pmstd{.0073} & .6543\pmstd{.0397} & 5.25\\
Graph token (no dropout) & .9122\pmstd{.0077} & \textbf{.7223}\pmstd{.0124} & .8356\pmstd{.0065} & .6403\pmstd{.0402} & 5.25\\
Graph token + dropout & .9123\pmstd{.0072} & .7096\pmstd{.0150} & .8378\pmstd{.0042} & .6558\pmstd{.0306} & 5.25\\
Masked logit stacking & .9217\pmstd{.0034} & .7069\pmstd{.0078} & .8309\pmstd{.0059} & \textbf{.6636}\pmstd{.0348} & 5.00\\
Naive residual & .9228\pmstd{.0046} & .7113\pmstd{.0070} & .8292\pmstd{.0136} & .6261\pmstd{.0435} & 6.25\\
TGRF-Frozen & \underline{.9233}\pmstd{.0050} & .7044\pmstd{.0069} & \textbf{.8457}\pmstd{.0046} & .6538\pmstd{.0329} & \underline{4.25}\\
TGRF-Joint & \textbf{.9238}\pmstd{.0038} & \underline{.7131}\pmstd{.0100} & \underline{.8451}\pmstd{.0122} & \underline{.6596}\pmstd{.0282} & \textbf{1.75}\\
\bottomrule
\end{tabular*}
\end{table*}

\textbf{Task-attention based expert fusion.}
\label{sec:fusion-evaluation}
We compare fusion methods with the upstream models fixed (Table~\ref{tab:fusion-main-a}). \textbf{Task-attention based expert fusion achieves the best AUROC, AP, and R@1\% on Sampled Industrial.} It also leads all three metrics on TransXion. Its R@1\% exceeds UMI, the strongest competing method, by 1.72 percentage points on Sampled Industrial (+5.44 vs.\ +3.72 over B1) and 1.11 points on TransXion (89.69\% vs.\ 88.58\%). The gains over the best competitors on IBM CCT and MBD are 8.09 and 3.33 points (Appendix~\ref{app:fusion-results}).

\begin{table*}[!htbp]
\caption{Task-conditioned fusion with fixed upstream models: Sampled Industrial (percentage-point difference from B1) and TransXion (\%). Mean $\pm$ standard deviation over five seeds.}
\label{tab:fusion-main-a}
\small
\begin{tabular*}{\textwidth}{@{\extracolsep{\fill}}lrrrrrr@{}}
\toprule
& \multicolumn{3}{c}{Sampled Industrial (pp vs.\ B1)} & \multicolumn{3}{c}{TransXion (\%)}\\
\cmidrule(lr){2-4}\cmidrule(lr){5-7}
Method & AUROC & AP & R@1\% & AUROC & AP & R@1\%\\
\midrule
B1 & +0.00 & +0.00 & +0.00 & 82.47 & 13.21 & 39.47\\
B2 & +0.85\pmstd{0.11} & +0.37\pmstd{0.35} & +1.77\pmstd{0.21} & 93.69\pmstd{0.07} & 18.85\pmstd{0.41} & 60.31\pmstd{0.52}\\
Concat & +1.78\pmstd{0.17} & +2.67\pmstd{0.22} & +1.99\pmstd{0.44} & 94.30\pmstd{0.07} & 18.76\pmstd{0.37} & 62.64\pmstd{0.62}\\
Sum-pooling & $-$0.67\pmstd{0.13} & +1.37\pmstd{0.31} & +1.81\pmstd{0.21} & 94.60\pmstd{0.08} & 19.26\pmstd{0.65} & 62.97\pmstd{0.96}\\
CE & $-$0.53\pmstd{0.14} & +1.19\pmstd{0.48} & +1.71\pmstd{0.71} & 94.83\pmstd{0.06} & 18.89\pmstd{0.82} & 62.75\pmstd{1.25}\\
MMT & $-$0.94\pmstd{0.25} & +1.67\pmstd{0.87} & +1.96\pmstd{1.30} & 94.63\pmstd{0.28} & 17.90\pmstd{1.00} & 61.64\pmstd{1.51}\\
UMI & +0.43\pmstd{0.11} & $-$6.35\pmstd{0.23} & +3.72\pmstd{0.27} & 98.59\pmstd{0.09} & 28.87\pmstd{0.63} & 88.58\pmstd{0.98}\\
$I^2M^2$ & $-$0.13\pmstd{0.21} & $-$3.07\pmstd{0.47} & +2.87\pmstd{0.65} & 99.10\pmstd{0.09} & 29.47\pmstd{0.34} & 87.14\pmstd{0.67}\\
MMPareto & +1.67\pmstd{0.17} & +0.37\pmstd{0.15} & +0.50\pmstd{0.34} & 93.91\pmstd{0.13} & 18.96\pmstd{0.47} & 62.20\pmstd{0.74}\\
$I^2$MoE & +1.29\pmstd{0.26} & +2.59\pmstd{0.68} & +1.48\pmstd{0.69} & 94.50\pmstd{0.09} & 18.67\pmstd{0.98} & 62.20\pmstd{0.94}\\
BALM & +1.46\pmstd{0.38} & +0.64\pmstd{0.54} & +1.27\pmstd{0.38} & 94.75\pmstd{0.08} & 18.97\pmstd{0.70} & 62.08\pmstd{0.64}\\
Task expert fusion & \textbf{+1.95}\pmstd{0.16} & \textbf{+3.10}\pmstd{0.28} & \textbf{+5.44}\pmstd{0.37} & \textbf{99.15}\pmstd{0.18} & \textbf{29.51}\pmstd{0.67} & \textbf{89.69}\pmstd{0.92}\\
\bottomrule
\end{tabular*}
\end{table*}

Table~\ref{tab:fusion-budgets-a} compares recall at intervention fractions of 0.5\%, 1\%, 5\%, 10\%, and 25\%. Task-attention based expert fusion leads at 0.5\% and 1\% on both datasets and at every displayed budget on Sampled Industrial. Appendix~\ref{app:fusion-results} gives the IBM CCT and MBD comparisons.

\begin{table}[!htbp]
\caption{Recall among the top-$q$ scored transactions: Sampled Industrial (pp vs.\ B1) and TransXion (\%).}
\label{tab:fusion-budgets-a}
\small
\setlength{\tabcolsep}{2pt}
\begin{tabular*}{\columnwidth}{@{\extracolsep{\fill}}lrrrrr@{}}
\toprule
\multicolumn{6}{@{}l}{\textit{Sampled Industrial (pp vs.\ B1)}}\\
Method / $q$ & 0.5\% & 1\% & 5\% & 10\% & 25\%\\
\midrule
B1 & +0.0 & +0.0 & +0.0 & +0.0 & +0.0\\
B2 & +1.8 & +1.8 & +1.4 & +0.7 & +0.0\\
Concat & +1.2 & +2.0 & +1.4 & +1.0 & +0.3\\
CE & +1.6 & +1.7 & $-$0.8 & $-$1.7 & $-$0.1\\
MMT & +2.1 & +2.0 & $-$1.2 & $-$1.7 & $-$0.3\\
$I^2M^2$ & +2.6 & +2.9 & +2.2 & +1.3 & +0.4\\
UMI & +0.6 & +3.7 & +1.8 & +0.9 & $-$0.7\\
\makecell[l]{Task expert\\fusion} & \textbf{+3.3} & \textbf{+5.4} & \textbf{+4.2} & \textbf{+3.0} & \textbf{+1.7}\\
\midrule
\multicolumn{6}{@{}l}{\textit{TransXion (\%)}}\\
Method / $q$ & 0.5\% & 1\% & 5\% & 10\% & 25\%\\
\midrule
B2 & 53.9 & 60.3 & 74.2 & 81.2 & 90.9\\
Concat & 53.2 & 62.6 & 78.1 & 85.0 & 92.7\\
CE & 54.8 & 62.8 & 77.8 & 86.5 & 93.2\\
MMT & 49.6 & 61.6 & 78.4 & 85.8 & 93.0\\
$I^2M^2$ & 78.3 & 87.1 & 96.9 & 97.9 & \textbf{99.3}\\
UMI & 78.0 & 88.6 & \textbf{97.0} & \textbf{98.2} & 98.2\\
\makecell[l]{Task expert\\fusion} & \textbf{80.9} & \textbf{89.7} & 96.9 & 98.1 & \textbf{99.3}\\
\bottomrule
\end{tabular*}
\end{table}

\textbf{Overall findings.}
These results demonstrate the effectiveness of the proposed methods across sequence pretraining, graph augmentation, and expert fusion. PFC improves downstream performance under limited supervision by learning informative sequence representations, TGRF introduces complementary relational information to strengthen risk prediction, and task-attention based expert fusion further improves detection performance through adaptive expert weighting, particularly at low intervention budgets. The consistent gains across diverse datasets, label budgets, and evaluation settings provide further evidence of the effectiveness and robustness of our proposed framework.

\subsection{Modality Ablation}
\label{sec:modality-ablation}

We evaluate all combinations of sequence (S), tabular (T), and graph (G) modalities(Table~\ref{tab:modality-ablation}). The full SGT model (\model{}) achieves the highest AP (3.89\%), R@0.5\% (10.08\%), and AddCB@0.5\% (7.59\%). Removing any modality reduces the reported mean performance, indicating the effectiveness of each modality. Sequence provides the strongest standalone signal (2.46\% AP), and removing it causes the largest performance drop (3.89\% to 1.06\% AP). Removing graph or tabular reduces AP to 2.94\% and 2.49\%, respectively.

\begin{table*}[!htbp]
\caption{Modality ablation on the industrial sampled offline dataset (\%).}
\label{tab:modality-ablation}
\small
\begin{tabular*}{\textwidth}{@{\extracolsep{\fill}}lrrrr@{}}
\toprule
Variant & AP & R@0.5\% & AddCB@0.5\%\\
\midrule
\textbf{Full SGT (\model{})} & \textbf{3.89 $\pm$ 0.40} & \textbf{10.08 $\pm$ 3.24} & \textbf{7.59 $\pm$ 1.64}\\
w/o Sequence: GT & 1.06 $\pm$ 0.26 & 1.79 $\pm$ 1.33 & 0.95 $\pm$ 1.09\\
w/o Graph: ST & 2.94 $\pm$ 0.19 & 5.53 $\pm$ 0.56 & 6.05 $\pm$ 1.98\\
w/o Tabular: SG & 2.49 $\pm$ 1.05 & 4.36 $\pm$ 1.35 & 4.63 $\pm$ 0.00\\
Sequence only: S & 2.46 $\pm$ 1.00 & 4.25 $\pm$ 1.22 & 4.51 $\pm$ 0.41\\
Tabular only: T & 0.81 $\pm$ 0.02 & 0.33 $\pm$ 0.10 & 0.00 $\pm$ 0.00\\
Graph only: G & 0.65 $\pm$ 0.13 & 0.72 $\pm$ 0.44 & 0.24 $\pm$ 0.41\\
\bottomrule
\vspace{2mm}
\end{tabular*}\end{table*}

\subsection{Case Study}
\label{sec:seq-behavior}

We inspect whether pretrained representations distinguish payment behaviors and how individual histories occupy the learned representation space.

\begin{figure}[!htbp]
\centering
  \ifReportHTML
  \includegraphics[width=\linewidth,alt={PCA and t-SNE projections of pretrained representations, with two behavior groups and an outlined t-SNE region used for case inspection.}]{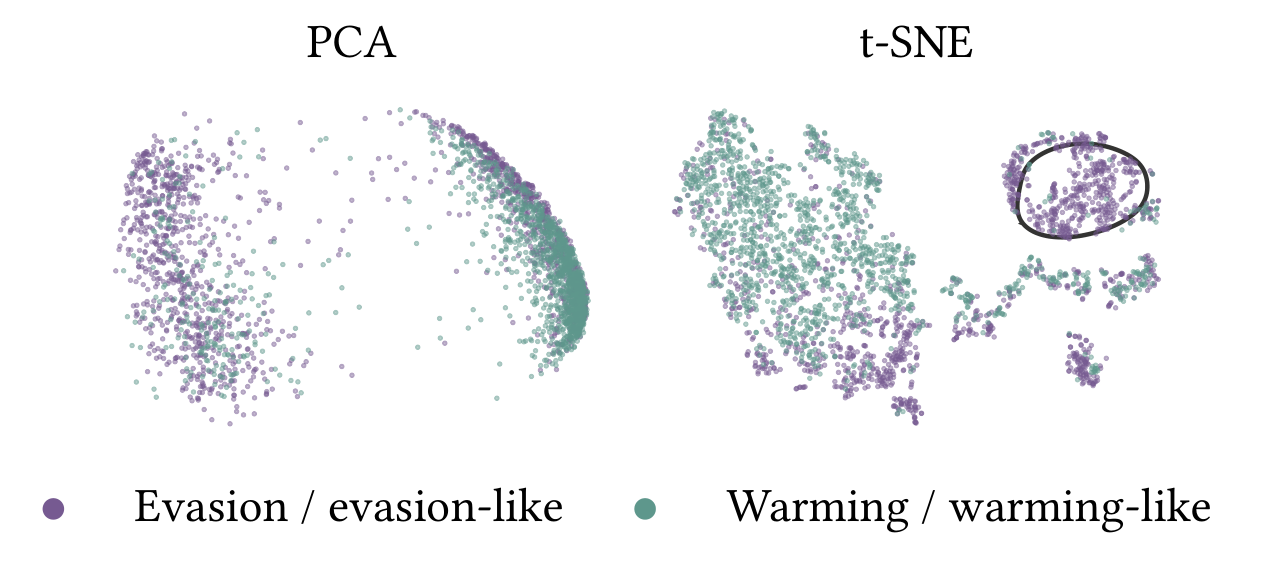}
  \else
  \incfig[\columnwidth]{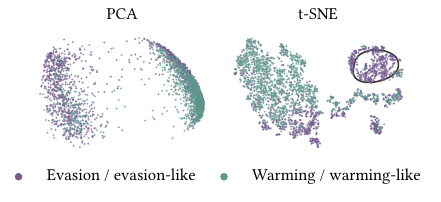}
  \fi
  \caption{Pretrained representations of 2,880 annotated histories: PCA (left) and t-SNE (right).}
  \Description{PCA and t-SNE projections of pretrained representations, with two behavior groups and an outlined t-SNE region used for case inspection.}
\label{fig:behavior-analysis}
\end{figure}

\textbf{Evasion and account nurturing.} Evasion changes payment resources or attempt timing to get past risk controls, while account nurturing builds routine-looking activity before later misuse. The pretrained representations of these two groups form two largely separated regions in PCA. We inspect 2,880 annotated histories, including both risk patterns and legitimate histories with similar behavior, and visualize their representations with PCA and t-SNE (Figure~\ref{fig:behavior-analysis}).
The region contains 374 histories and is dominated by evasion-type histories.

\begin{figure}[!htbp]
\centering
  \ifReportHTML
  \includegraphics[width=\linewidth,alt={Three payment attempts show changes in anonymous card and device identities following rejections while retaining the IP identity, ending in a successful attempt.}]{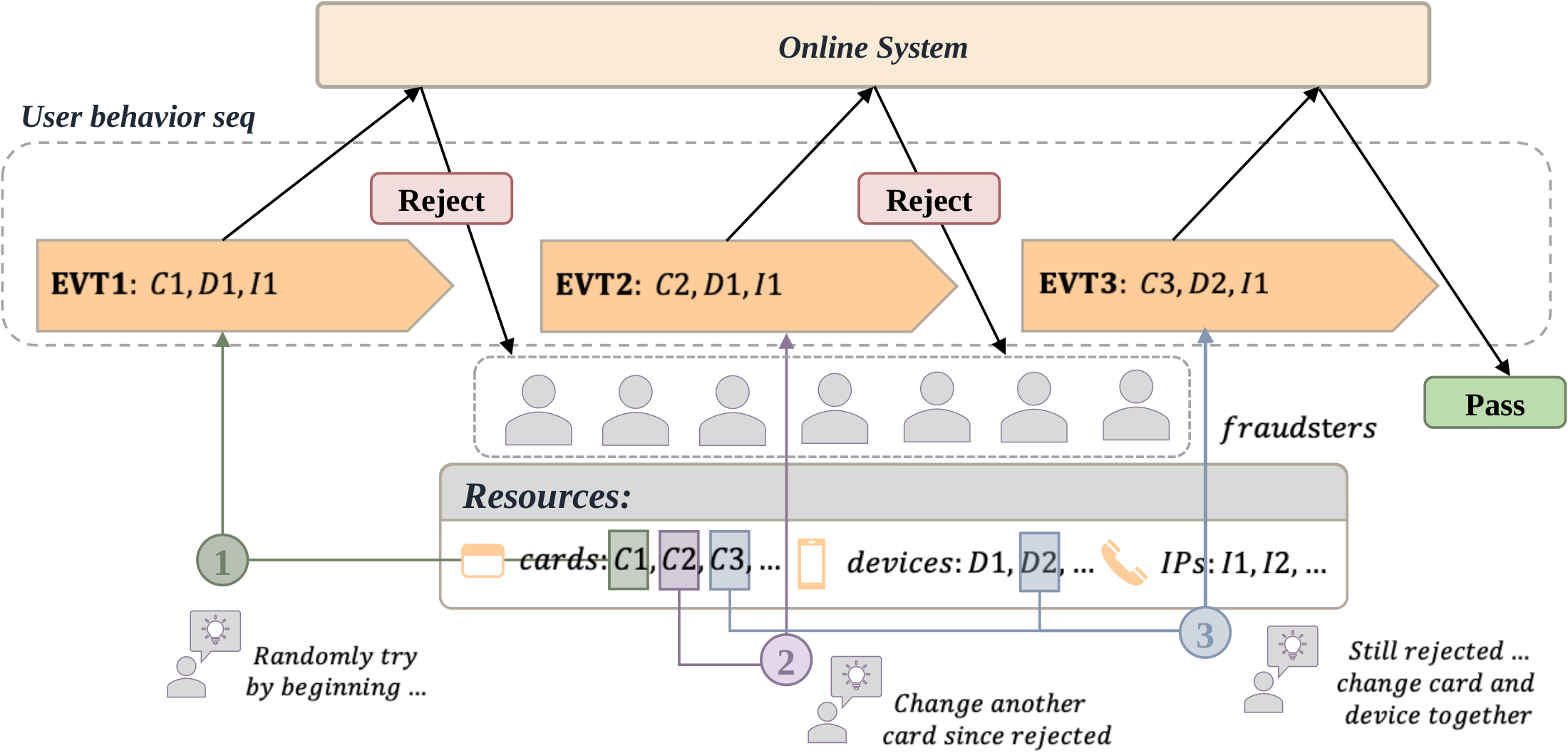}
  \else
  \incfig[\columnwidth]{figs/fig_case_retry.pdf}
  \fi
  \caption{Card and device switching after rejection, from the outlined region in Figure~\ref{fig:behavior-analysis}.}
  \Description{Three payment attempts show changes in anonymous card and device identities following rejections while retaining the IP identity, ending in a successful attempt.}
  \label{fig:behavior-cases}
\end{figure}

\textbf{Behavior pattern case study.} Samples in the outlined region of Figure~\ref{fig:behavior-analysis} show a typical behavior pattern, rapidly retrying at the same merchant. Figure~\ref{fig:behavior-cases} details the pattern. Among 361 annotated feedback--adjustment pairs in the region, 95.6\% retain the same merchant, compared with 70.7\% among 916 pairs outside the surrounding island. Median intervals are 4.8 minutes and 22.8 hours, respectively. After a rejection, the example changes the card while retaining the device and IP address. After another rejection, both card and device change while the IP address stays. The third attempt succeeds 120 seconds after the first.

\section{Related Work}
\label{sec:related}
\label{app:related}
\subsection{Transaction Risk Modeling}

Transaction risk models have moved from hand-crafted features toward sequential and graph-based learning.
Early systems feed manually designed statistics, such as velocity features, into tabular learners such as gradient-boosted trees~\cite{chen2016xgboost,prokhorenkova2018catboost}.
Later work models transaction histories directly with sequential networks~\cite{jurgovsky2018sequence,zhang2023fatatrans}, or connects accounts, devices, and merchants into graphs and applies graph neural networks~\cite{liu2018heterogeneous,wang2019semi,dou2020caregnn,liu2021pcgnn,tang2022bwgnn}.
Recent advances in large language models have
spurred the development of transaction foundation
models~\cite{ref1,ref2}.

\subsection{Pretrained Transaction Foundation Models}
Transaction pretraining uses two main forms of supervision: event reconstruction learns dependencies among heterogeneous fields through masking or next-event prediction, as in TabBERT, FATA-Trans, TransactionGPT, and PRAGMA~\cite{padhi2021tabformer,zhang2023fatatrans,ref1,ref2}; temporal learning instead relates segments of a history: CoLES contrasts subsequences, while CPC and BYB predict future representations~\cite{babaev2022coles,oord2018cpc,wu2025byb}. Transaction foundation models extend these objectives with profile information and reusable representations for multiple business tasks~\cite{ref3,zhao2026id2pattern}.

Despite their achievements in transaction modeling, they mainly capture transaction features from a single perspective. \model{} uses three modalities to provide a more comprehensive view of transactions.

\section{Conclusion}
\label{sec:conclusion}

We presented \model{}, a payment foundation model applied on billion-scale industrial payment systems, which combines transaction sequences, scenario-specific tabular features, and entity relationships for card-payment risk. Sequence modality captures behavioral evolution, tabular modality models contextual feature interactions, and graph modality learns structure of user relationships. \model{} has been deployed on Ant International's industrial system since June 2026 and reduces overall loss rate by 37.88\% relative to current internal fraud detection system. \model{} demonstrates promising generalization capabilities across diverse scenarios and tasks. The technologies developed in this work have also strengthened Ant International's overall risk management capabilities.

\FloatBarrier
\ifReportHTML\else\renewcommand{\bibfont}{\normalfont\fontsize{9}{11.5}\selectfont}\fi
\ifReportHTML\else

\appendix
\section{Metrics and Comparison Conventions}
\label{app:metrics}

\subsection{Risk Ranking at a Fixed Budget}

Let $s_i$ be the risk score, $y_i\in\{0,1\}$ the confirmed label, and $w_i$ the population weight of payment $i$. A complete evaluation population uses $w_i=1$; a sampled population uses inverse-inclusion weights. At intervention fraction $q$, let $a_i(q)\in[0,1]$ indicate the fraction of payment $i$ selected by the score ranking. Nonboundary payments have $a_i(q)\in\{0,1\}$. In the sampled evaluations, payments tied at the budget boundary share a fractional allocation satisfying
\begin{equation}
\sum_i a_i(q)w_i=q\sum_iw_i,\qquad
\mathrm{R@}q=\frac{\sum_i a_i(q)w_i y_i}{\sum_iw_i y_i}.
\label{eq:recall}
\end{equation}
This keeps the selected population weight fixed when identical histories receive identical scores.

Average precision integrates precision over the empirical recall increments:
\begin{equation}
\mathrm{AP}=\sum_j (R_j-R_{j-1})P_j,
\label{eq:ap}
\end{equation}
where $P_j$ and $R_j$ are the weighted precision and recall after including the $j$-th score group. AP is computed without interpolation using the weighted precision--recall curve. AUROC evaluates ranking over all positive--negative pairs; R@$q$ evaluates the portion of that ranking used for intervention. The public benchmark's F1 threshold is selected on validation data and then fixed.

\subsection{Business Metrics and Normalized Results}

Some business studies report an anonymized operating metric $M$ against a reference $M_0$. We retain the metric and reference associated with each study. A relative gain and a normalized index are, respectively,
\begin{equation}
\Delta_{\mathrm{rel}}=100\left(\frac{M}{M_0}-1\right)\%,\qquad
I=100\frac{M}{M_0}.
\label{eq:relative-gain}
\end{equation}
An index of 120 therefore denotes a relative gain of 20\%. A difference between percentage-valued rates is reported in percentage points. Absolute AUROC changes are labeled as absolute differences.

The source-domain transfer study compares a pretrained encoder with scratch training on the same target-business labels. Its reported business-metric changes are +34.5\% for the full cohort and +34.2\% after removing high-concentration accounts. The joint-model study compares models at the same intervention volume and reports gains over the baseline. These references remain separate throughout the report.

Business metrics are reported as within-study normalized indices or relative gains. The tables identify their reference models; the joint-model comparison additionally fixes the intervention volume. AP, fixed-budget recall, and AUROC retain their explicit definitions above.

\subsection{Additional Captured Risk Events(AddCB)}

For the joint-model comparison and modality ablation, let $A_q$ and $B_q$ be the payment sets selected by the developed model and existing baselines at the same volume. The share of captured risk events missed by the baseline is
\begin{equation}
\mathrm{AdditionalCoverage}(q)=
\frac{\sum_{i\in A_q\setminus B_q}y_i}{\sum_{i\in A_q}y_i}.
\label{eq:additional-coverage}
\end{equation}
It describes the share of its captured risk events outside the baseline's selection.

\section{Supplementary Evaluation}
\label{app:supplementary-evaluation}

\subsection{Dataset Details}
\label{app:datasets}
Table~\ref{tab:seq-data} gives the low-label and cross-scenario dataset statistics, Table~\ref{tab:graph-protocol} the graph benchmark statistics, and Table~\ref{tab:fusion-data} the fusion dataset splits.

Sampled Industrial contains industrial card payments with tabular features, user/card/device sequences, and user graphs. TransXion~\cite{chen2026transxion} is an anti-money-laundering benchmark with current events, sender and receiver histories, and a graph of prior events. IBM CCT uses IBM credit-card transactions with tabular features, user and card histories, and a prior-event graph under a separate split. The Multimodal Banking Dataset (MBD)~\cite{mollaev2025mbd} predicts four next-month product purchases from client profiles and transaction summaries.

\begin{table}[!htbp]
\caption{Industrial datasets for the low-label and cross-scenario studies.}
\label{tab:seq-data}
\small
\begin{tabular*}{\columnwidth}{@{\extracolsep{\fill}}lrrrr@{}}
\toprule
Domain & \makecell[r]{Scored\\payments} & \makecell[r]{History\\events} & \makecell[r]{Mean\\length} & \makecell[r]{Empty\\history}\\
\midrule
\merchantA{} & 501,898 & 14.78M & 29.5 & 30.63\%\\
\merchantB{} & 371,600 & 2.91M & 7.8 & 76.91\%\\
HK & 127,375 & 16.03M & 125.8 & 0.50\%\\
\bottomrule
\end{tabular*}
\end{table}

\begin{table}[!htbp]
\caption{Graph benchmark statistics~\cite{tang2023gadbench}.}
\label{tab:graph-protocol}
\small
\setlength{\tabcolsep}{3pt}
\begin{tabular*}{\columnwidth}{@{\extracolsep{\fill}}lrrrrr@{}}
\toprule
Dataset & Nodes & Edges & Features & Relations & Anomaly\\
\midrule
Amazon & 11,944 & 4,398,392 & 25 & 3 & 9.5\%\\
YelpChi & 45,954 & 3,846,979 & 32 & 3 & 14.5\%\\
T-Finance & 39,357 & 21,222,543 & 10 & 1 & 4.6\%\\
Elliptic & 203,769 & 234,355 & 166 & 1 & 9.8\%\\
\bottomrule
\end{tabular*}
\end{table}

\begin{table}[!htbp]
\caption{Datasets for the fusion studies: transactions, or client-months for MBD.}
\label{tab:fusion-data}
\small
\setlength{\tabcolsep}{2pt}
\begin{tabular*}{\columnwidth}{@{\extracolsep{\fill}}lrrr@{}}
\toprule
Dataset & Train & Validation & Test\\
\midrule
Sampled Industrial & 5,832,600 & 1,392,465 & 1,366,237\\
TransXion & 1,817,503 & 605,834 & 605,833\\
IBM CCT & 19,512,454 & 2,436,496 & 2,437,950\\
MBD & 801,792 & 200,448 & 200,448\\
\bottomrule
\end{tabular*}
\end{table}

\subsection{Online Evaluation by Transaction Date}
\label{app:online-stages}
The two-stage evaluation in Table~\ref{tab:online-stages} uses card-payment traffic from the Antom platform. Stage~I evaluates scores without model-triggered interventions. Stage~II evaluates production decisions. The Stage~II chargeback snapshot uses 66 common transaction dates with recorded chargeback amounts and positive successful-transaction amounts in all arms, as observed on October 25, 2026. Pooled rates divide total chargeback amount by total successful amount. Date-weighted rates average each arm's daily chargeback rates using the combined successful amounts across arms as common weights. Portfolio results in Table~\ref{tab:online-ab} pool the evaluated payment businesses within each arm and attribute chargebacks to their original transaction dates. Traffic shares shifted briefly during load testing and staged rollout, while assignment stayed random.

\subsection{Additional Fusion Datasets}
\label{app:fusion-results}
Task-conditioned fusion is also evaluated on IBM CCT and MBD. Table~\ref{tab:fusion-main-b} gives the
main results.

\begin{table*}[!htbp]
\caption{Task-conditioned fusion with fixed upstream models: IBM CCT (\%) and MBD (\%). Mean $\pm$ standard deviation over five seeds.}
\label{tab:fusion-main-b}
\small
\begin{tabular*}{\textwidth}{@{\extracolsep{\fill}}lrrrrrr@{}}
\toprule
& \multicolumn{3}{c}{IBM CCT (\%)} & \multicolumn{3}{c}{MBD (\%)}\\
\cmidrule(lr){2-4}\cmidrule(lr){5-7}
Method & AUROC & AP & R@1\% & AUROC & AP & R@1\%\\
\midrule
B1 & 94.06 & 3.06 & 34.01 & 68.20 & 0.396 & 3.18\\
B2 & 95.33\pmstd{0.07} & 4.09\pmstd{0.25} & 36.51\pmstd{0.46} & 57.62\pmstd{0.42} & 0.308\pmstd{0.009} & 2.63\pmstd{0.17}\\
Concat & 94.69\pmstd{0.40} & 4.69\pmstd{1.07} & 36.25\pmstd{1.75} & 72.84\pmstd{0.81} & 0.647\pmstd{0.024} & 6.20\pmstd{0.14}\\
Sum-pooling & 94.62\pmstd{0.13} & 4.54\pmstd{0.28} & 36.07\pmstd{0.35} & 71.55\pmstd{0.73} & 0.605\pmstd{0.032} & 5.97\pmstd{0.24}\\
CE & 94.85\pmstd{0.30} & 4.74\pmstd{0.61} & 36.84\pmstd{1.39} & 72.38\pmstd{0.39} & 0.530\pmstd{0.013} & 4.48\pmstd{0.43}\\
MMT & 95.29\pmstd{0.66} & 5.11\pmstd{1.06} & 39.34\pmstd{1.82} & 72.25\pmstd{0.36} & 0.533\pmstd{0.033} & 5.32\pmstd{0.52}\\
UMI & 92.93\pmstd{0.21} & 12.16\pmstd{0.27} & 53.37\pmstd{0.92} & 72.35\pmstd{0.32} & 0.562\pmstd{0.062} & 5.28\pmstd{0.39}\\
$I^2M^2$ & 94.92\pmstd{0.30} & 10.88\pmstd{0.39} & 51.27\pmstd{1.36} & 73.24\pmstd{0.75} & 0.596\pmstd{0.016} & 4.76\pmstd{0.26}\\
MMPareto & 94.58\pmstd{0.18} & 4.78\pmstd{0.71} & 36.69\pmstd{0.79} & 72.76\pmstd{0.47} & 0.596\pmstd{0.036} & 6.36\pmstd{0.54}\\
$I^2$MoE & 94.93\pmstd{0.49} & 5.02\pmstd{0.78} & 37.43\pmstd{2.49} & 59.68\pmstd{0.53} & 0.418\pmstd{0.020} & 2.60\pmstd{0.51}\\
BALM & 94.87\pmstd{0.30} & 5.48\pmstd{0.82} & 38.31\pmstd{1.32} & 72.94\pmstd{0.74} & 0.590\pmstd{0.040} & 6.87\pmstd{0.67}\\
Task-conditioned fusion & \textbf{96.82}\pmstd{0.23} & \textbf{22.36}\pmstd{0.29} & \textbf{61.46}\pmstd{1.46} & \textbf{77.80}\pmstd{0.90} & \textbf{0.677}\pmstd{0.033} & \textbf{10.20}\pmstd{0.62}\\
\bottomrule
\end{tabular*}
\end{table*}

The learned fusion references include CE~\cite{liu2019ce}, MMT~\cite{gabeur2020mmt}, UMI~\cite{zhang2023umi}, $I^2M^2$~\cite{madaan2024i2m2}, MMPareto~\cite{wei2024mmpareto}, $I^2$MoE~\cite{xin2025i2moe}, and BALM~\cite{nguyen2026balm}. They receive the same fixed upstream outputs within each dataset. B1 is the fixed single-model reference, while B2 combines prediction scores.

On IBM CCT, task-conditioned fusion raises R@1\% to 61.46\%, compared with 53.37\% for UMI, and AP to 22.36\% from 12.16\%. On MBD, it reaches 10.20\% R@1\%, compared with 6.87\% for BALM. MBD averages four product-purchase targets equally; these values measure customer targeting rather than fraud detection.

\begin{table}[!htbp]
\caption{Recall among the top-$q$ scored transactions: IBM CCT and MBD (\%).}
\label{tab:fusion-budgets-b}
\small
\setlength{\tabcolsep}{2pt}
\begin{tabular*}{\columnwidth}{@{\extracolsep{\fill}}lrrrrr@{}}
\toprule
\multicolumn{6}{@{}l}{\textit{IBM CCT (\%)}}\\
Method / $q$ & 0.5\% & 1\% & 5\% & 10\% & 25\%\\
\midrule
B2 & 23.0 & 36.5 & 73.8 & 87.7 & 96.7\\
Concat & 22.0 & 36.3 & 71.7 & 86.1 & 95.6\\
CE & 23.2 & 36.8 & 72.7 & 86.6 & 95.6\\
MMT & 23.0 & 39.3 & 75.7 & 88.2 & 96.0\\
$I^2M^2$ & 37.4 & 51.3 & 80.7 & 89.3 & 94.6\\
UMI & 39.7 & 53.4 & \textbf{83.7} & 89.5 & 91.1\\
\makecell[l]{Task-conditioned\\fusion} & \textbf{50.3} & \textbf{61.5} & \textbf{83.7} & \textbf{91.4} & \textbf{97.2}\\
\midrule
\multicolumn{6}{@{}l}{\textit{MBD (\%)}}\\
Method / $q$ & 0.5\% & 1\% & 5\% & 10\% & 25\%\\
\midrule
B2 & 1.7 & 2.6 & 9.8 & 17.6 & 36.2\\
Concat & 3.6 & 6.2 & 20.2 & 33.4 & 58.0\\
CE & 1.9 & 4.5 & 18.5 & 30.7 & 56.1\\
MMT & 2.4 & 5.3 & 18.5 & 32.1 & 55.5\\
$I^2M^2$ & 2.9 & 4.8 & \textbf{22.5} & 33.0 & 56.7\\
BALM & 4.4 & 6.9 & 22.2 & 32.8 & 58.0\\
\makecell[l]{Task-conditioned\\fusion} & \textbf{7.7} & \textbf{10.2} & 21.7 & \textbf{35.2} & \textbf{64.4}\\
\bottomrule
\end{tabular*}
\end{table}

\end{document}